\RequirePackage[hyphens]{url}
\documentclass{article}

\usepackage{microtype}
\usepackage{graphicx}
\usepackage{subfigure}
\usepackage{booktabs} 
\usepackage{natbib}
\usepackage{url}
\usepackage{hyperref}
\usepackage{fancyhdr}

\usepackage[accepted]{icml2021}

\icmltitlerunning{Meat Freshness Prediction}
\fancypagestyle{plain}{}
\begin{document}

\twocolumn[
\icmltitle{Meat Freshness Prediction}


\begin{icmlauthorlist}
\icmlauthor{Bhargav Sagiraju}{nus}
\icmlauthor{Nathan Casanova}{nus}
\icmlauthor{Lam Ivan Chuen Chun}{nus}\\
\icmlauthor{Manan Lohia}{nus}
\icmlauthor{Toshinori Yoshiyasu}{nus}
\end{icmlauthorlist}

\icmlaffiliation{nus}{Business Analytics Centre, National University of Singapore, Singapore, Singapore}

\icmlcorrespondingauthor{Bhargav Sagiraju}{bhargav.sagiraju@u.nus.edu}

\vspace{0.09in}

\vskip 0.3in
]



\printAffiliationsAndNotice{} 

\begin{abstract}
In most retail stores, the number of days since initial processing is used as a proxy for estimating the freshness of perishable foods or freshness is assessed manually by an employee. While the former method can lead to wastage, as some fresh foods might get disposed after a fixed number of days, the latter can be time-consuming, expensive and impractical at scale. This project aims to propose a Machine Learning (ML) based approach that evaluates freshness of food based on live data. For the current scope, it only considers meat as a the subject of analysis and attempts to classify pieces of meat as fresh, half-fresh or spoiled. Finally the model achieved an accuracy of above 90\% and relatively high performance in terms of the cost of misclassification. It is expected that the technology will contribute to the optimization of the client's business operation, reducing the risk of selling defective or rotten products that can entail serious monetary, non-monetary and health-based consequences while also achieving higher corporate value as a sustainable company by reducing food wastage through timely sales and disposal.
\end{abstract}

\section{Business Problem}
\label{Business Problem}

Assessing the freshness of perishable food is a significant operational challenge for retailers, as it is time-consuming, and can affect their business performance as well as reputation if a wrong judgment is made. In most retail stores, the number of days since initial processing is used as a proxy for freshness. Regardless of actual freshness, products are judged to be fresh if fewer days have passed and stale if more days have passed. Products that have passed many days since their initial processing are discounted, and if they are still not purchased, they are ultimately disposed of. However, freshness is a dynamic factor that depends on the product's processing and display environment. In other words, when freshness is uniformly judged based on the number of days since initial processing, fresh food may be discounted or discarded, while spoiled products may still be displayed. 

This project assumes a supermarket that sells fresh food including meat as the client and aims to create a system that evaluates freshness of meat based on actual meat conditions using images of meat as input. The system is expected to contribute to the optimization of the client’s business by avoiding unnecessary discount or disposal, help it reduce the risk of damage caused by selling defective products, and improve its corporate value as a sustainable company if they can reduce food waste.

\section{Assumption}
\label{Assumption}
This project sets some assumptions for the client’s business operation and customers’ behaviors.
\begin{itemize}
\item There are three classes in meat freshness: (1) Fresh (FS), (2) Half-fresh (HF), (3) Spoiled (SP)
\item Meat will be treated based on its “predicted” freshness as follows: (1) FS: Sold for \$10, (2) HF: Sold for \$5 (discounted), (3) SP: Discarded
\item The purchase probability of the meat depends on its “actual” freshness, which customer can tell based on its look, and price that is decided based on its “predicted” freshness. Actual FR has the purchase probability of 90\% when sold for the original price of \$10. It will be purchased with 100\% probability when sold for \$5, which is misclassification of actual FR being predicted as HF. The purchase probability of actual HF is just 10\% when sold for \$10, which is a misclassification of actual HF being predicted as FR, because of its less appetizing look. If it is discounted to \$5 based on the correct prediction, the probability increases to 90\%. For the same reason above, the purchase probability of actual SP is just 1\% and 5\% when sold for \$10 and \$5 respectively, which are the results of misclassification. Table \ref{Purchase Probability} shows the summary of purchase probabilities. The values in red indicate “misclassification”.

\begin{table}[ht]
\caption{Purchase Probability of Each Actual\textbar Price(Pred) Combination }
\label{Purchase Probability}
\vskip 0.05 in
\begin{center}
\begin{small}
\begin{sc}
\begin{tabular}{lcccr}
\toprule
Actual & \$10 (Pred as FR) & \$5 (Pred as HF) \\
\midrule
FR & 90\% & \textcolor{red}{100\%} \\
HF & \textcolor{red}{10\%} & 90\% \\
SP & \textcolor{red}{1\%} & \textcolor{red}{5\%} \\
\bottomrule
\end{tabular}
\end{sc}
\end{small}
\end{center}
\vskip -0.05in
\end{table}

\item If meat is purchased, it will be consumed and thus if spoiled meat is purchased, the customer will have a health issue. A total cost of \$100,000 will be incurred due to legal action, reporting to health-related authorities, loss of corporate trust, and other related factors.

\end{itemize}

\section{Methodology}
\label{Methodology}
This project experiments with two different models, namely ResNet and UNet to predict meat freshness, comparing their performance on a fixed set of metrics to identify the best performer. The original train dataset is split into train and validation dataset that are used for model development and hyperparameter tuning. The original validation set is treated as an ‘unseen’ test dataset that is only used for final model evaluation after all tuning has been done.

In evaluating model performance, a metric called MisClassification Cost (MCC), which is defined specifically for this project, is used. MCC is an expected value and represents the cost associated with misclassification, which depends on the actual class and the direction of the misclassification. A special metric apart from class based accuracy, precision, recall etc is required as the cost of misclassification is not symmetric, and hence certain misclassifications are more expensive or riskier than others. 

MCC is calculated by \textit{expected loss from misclassification} less \textit{expected gain from misclassification}, which considers the probability of purchase in each case explained in \ref{Assumption}.Assumption. Misclassification on actual SP samples leads to serious consequences due to customers' potential health issue it could cause. Furthermore, "misclassifiyng actual SP as HF" has the higher expected cost than "misclassifiyng actual SP as FR". This is because discounts due to "predicted HF" can increase the purchase probability, which increases the risk. Thus, predicting actual SP as HF is considered the most costly misclassification that should be avoided in this project. Table \ref{MCC} and \ref{Calculation of MCC} show MCC of each combination of actual and predicted classes and calculation of MCC.
\newpage

\begin{table}[ht]
\caption{MCC of Each Actual\textbar Pred Combination}
\label{MCC}
\vskip 0.05 in
\begin{center}
\begin{small}
\begin{sc}
\begin{tabular}{lcccr}
\toprule
Actual\textbar Pred & Consequence & MCC \\
\midrule
FR\textbar FR & Nothing(Correct) & \$0.0 \\
FR\textbar HF & Unnecessary discount & \$4.0 \\
FR\textbar SP & Unnecessary disposal & \$9.0 \\
HF\textbar HF & Nothing(Correct) & \$0.0 \\
HF\textbar FR & Inefficient pricing & \$3.5 \\
HF\textbar SP & Unnecessary disposal & \$4.5 \\
SP\textbar SP & Nothing(Correct) & \$0.0 \\
SP\textbar FR & Cost of \$100k if purchased & \$99.9 \\
SP\textbar HF & Cost of \$100k if purchased & \$499.8 \\
\bottomrule
\end{tabular}
\end{sc}
\end{small}
\end{center}
\vskip -0.05in
\end{table}
\begin{table}[ht]
\caption{Calculation of MCC}
\label{Calculation of MCC}
\vskip 0.05 in
\begin{center}
\begin{small}
\begin{sc}
\begin{tabular}{lcccr}
\toprule
Actual\textbar Pred & Expected Loss & Expected Gain \\
\midrule
FR\textbar FR & \$0 & \$0\\
FR\textbar HF & \$10*90\% & \$5*100\%\\
FR\textbar SP & \$10*90\% & \$0\\
HF\textbar HF & \$0 & \$0\\
HF\textbar FR & \$5*90\% & \$10*10\%\\
HF\textbar SP & \$5*90\% & \$0\\
SP\textbar SP & \$0 & \$0 \\
SP\textbar FR & \$10,000*1\% & \$10*1\%\\
SP\textbar HF & \$10,000*5\% & \$5*5\%\\
\bottomrule
\end{tabular}
\end{sc}
\end{small}
\end{center}
\vskip -0.05in
\end{table}

An ideal model is one which minimizes the MCC. Hyperparameter settings and corresponding total MCC are recorded in Weight and Biases in order to compare the performance of different models more easily. It has to be noted that the evaluation in this project highly depends on the assumptions set in the previous section. In the real business setting, MCC must be modified following the user's actual discounting policy, purchase probabilities and estimated cost of any consequences.

Simultaneously, to further enhance the model robustness, MCC was not used during training as an evaluation metric or the loss function. By doing this, the model is prevented from learning to optimize based on MCC and attempts to optimize on a different metric instead, namely the cross entropy loss. The benefit is that model evaluation is done independently of model training, avoiding any data leakage and increasing the reliability of the model on unseen data.

\section{Data Description \& EDA}
\label{Data Description & EDA}
This project assumes the Meat Freshness Image Dataset \cite{Vinayakshanawad2020} is the dataset provided by the client. The data consists of two folders for train and test datasets. The images are 416 x 416 pixels with the train dataset having 1,816 images and test dataset having 452 images. All images are of red meats.

There are three classes of images as discussed in Assumptions section: fresh, half-fresh and spoiled.
\begin{figure}[ht]
    \centering
    \includegraphics[scale=0.42]{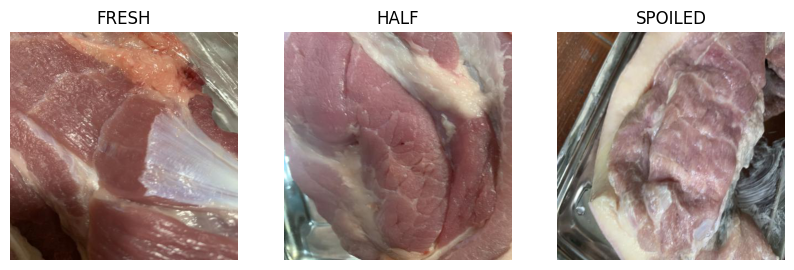}
    \caption{Sample Image per Class}
    \label{fig:Sample Image Per Class}
\end{figure}

During EDA, the class balance and pixel value frequency was explored to see if there was any abnormalities with the dataset before pre-processing. For class balance, the three classes were relatively balanced within the training dataset as shown below.

\begin{figure}[ht]
    \centering
    \includegraphics[scale=0.5]{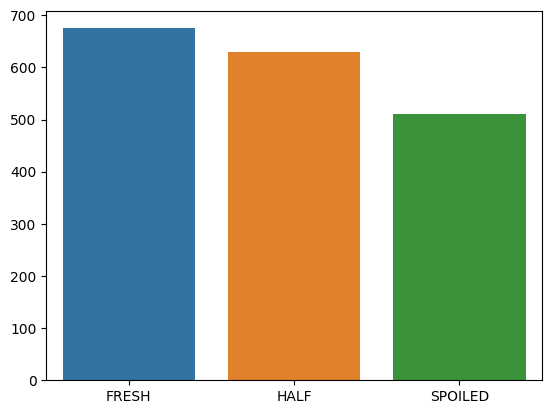}
    \caption{Number of Images per Class}
    \label{fig:Number of Images per Class}
\end{figure}

For pixel value frequency, it can be determined that the three classes have distinct distribution of pixel value frequencies as shown below.
\begin{figure}[!ht]
    \centering
    \includegraphics[scale=0.5]{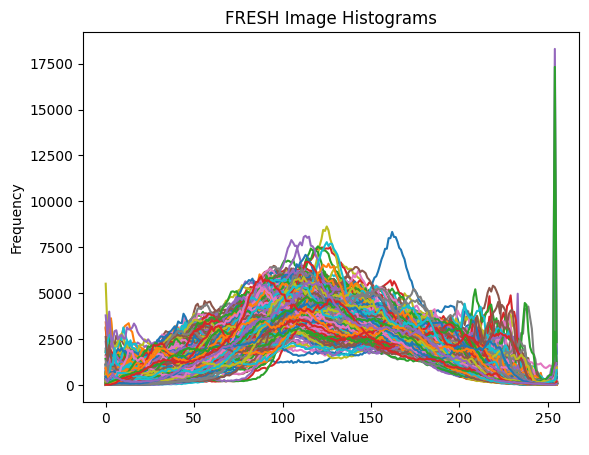}
    \caption{Pixel Value Distribution for Images of Fresh Meat. Pixel value from darker (0) to lighter (255).}
    \label{fig:Pixel Value Distribution for Images of Fresh Meat}
\end{figure}
\begin{figure}[!ht]
    \centering
    \includegraphics[scale=0.5]{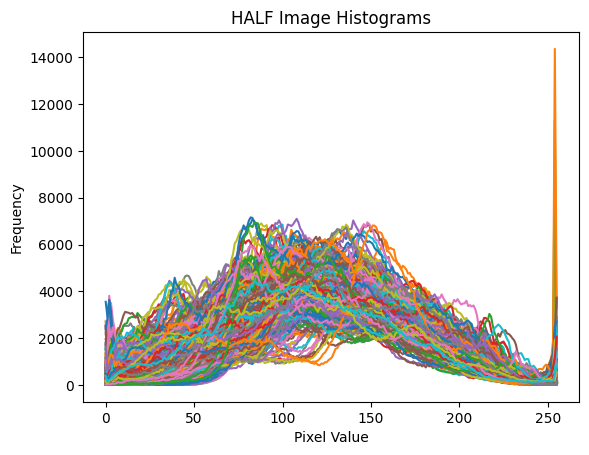}
    \caption{Pixel Value Distribution for Images of Half-Fresh Meat. Pixel value from darker (0) to lighter (255).}
    \label{fig:Pixel Value Distribution for Images of Fresh Meat}
\end{figure}
\newpage

\begin{figure}[!ht]
    \centering
    \includegraphics[scale=0.5]{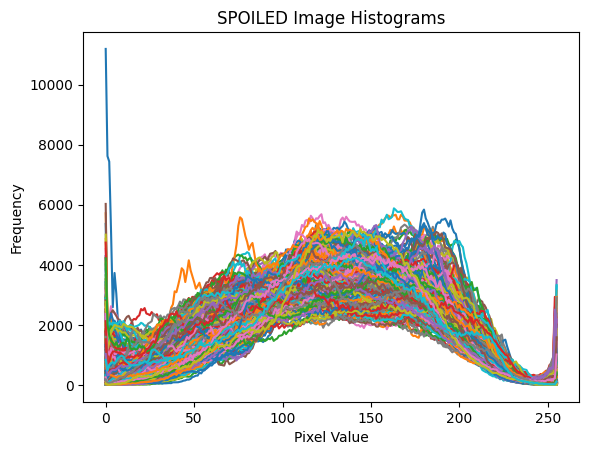}
    \caption{Pixel Value Distribution for Images of Spoiled Meat. Pixel value from darker (0) to lighter (255).}
    \label{fig:Pixel Value Distribution for Images of Fresh Meat}
\end{figure}

The distribution is significantly concentrated on the lighter pixels (255) for fresh meats while the distribution is concentrated on the dark pixels (0) for the spoiled meats. Half-fresh meat also has a distribution that is more concentrated on the lighter pixels than the spoiled meats. This is reasonable since on most meats the first sign of rot can be visibility detected by darker colored areas on the meat.

\section{Preprocessing }
\label{Preprocessing }

For pre-processing, augmentation was applied to the dataset in order for the models to be trained on more data with noise and also not overfit. The training dataset was split into training and validation and a class and pipeline transformation function was established to augment the dataset, the transformation function did the following on the training dataset:

\begin{enumerate}
    \itemsep0em
    \item Random Resized Crop
    \item Random Horizontal Flip
    \item Color Jitter
    \item Random Rotation
    \item Normalize per standard (mean=[0.485, 0.456, 0.406], std=[0.229, 0.224, 0.225])
\end{enumerate}

For the validation dataset, only standard augmentation was done:
\begin{enumerate}
    \itemsep0em
    \item Resize
    \item Center Crop
    \item Normalize per standard (mean=[0.485, 0.456, 0.406], std=[0.229, 0.224, 0.225])
\end{enumerate}

\begin{figure}[!ht]
    \centering
    \includegraphics[scale=0.5]{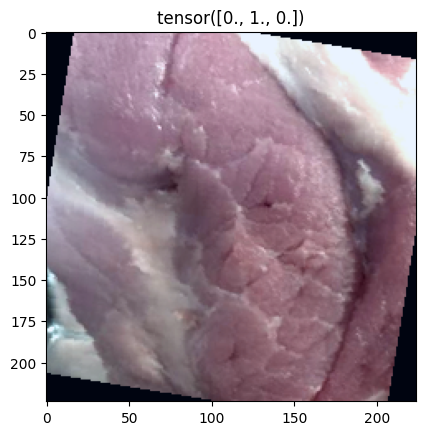}
    \caption{Sample Image after Transformation}
    \label{fig:Sample Image after Transformatio}
\end{figure}

\section{Algorithm and Modeling}
\label{Algorithm and Modeling}
\subsection{ResNet}
One of the models utilized to predict the freshness of the meat based on its image is ResNet \cite{ResNet}. ResNet is a convolutional neural network architecture that introduced residual blocks which allowed for effective training of deep neural networks. To fit ResNet for this paper's freshness prediction task, the final layer is reshaped to have the same output count as our image classes.

Two variants of the ResNet architecture were used to train models which are ResNet-18 and ResNet-50. The former is 18 layers deep while the latter is 50 layers deep. For model training, two transfer learning strategies were utilized. The first is feature extraction which utilized pre-trained weights of ResNet to get the image embeddings, and only the parameters of the final layer were updated during training. The pre-trained weights used for feature extraction were acquired from the PyTorch library \cite{PyTorch}. The second strategy is to fine-tune the whole ResNet architecture by updating all model parameters using the dataset. After training, the models were evaluated on the test set and the models' performance metrics and final hyperparameters used are reported in Tables \ref{ResNet Model Performance} and \ref{Resnet Final Hyperparameters} below. All model weights are saved after training should the client decide that the solution proposed is suitable for deployment.

\begin{table}[ht]
\caption{ResNet Model Performance \\FE: Feature Extraction, FT: Fine-tuning}
\centering
\label{ResNet Model Performance}
\vskip 0.05 in
\begin{center}
\begin{small}
\begin{sc}
\begin{tabular}{lcccr}
\toprule
Model & Accuracy & Precision & Recall \\
\midrule
ResNet 18-FE & 82.71\% & 84.66\% & 83.71\% \\
ResNet 18-FT  & \textbf{93.13\%} & \textbf{93.97\%} & \textbf{92.85\%} \\
ResNet 50-FE   & 88.03\% & 88.35\% & 88.86\% \\
ResNet 50-FT   & 84.70\% & 84.50\% & 86.14\% \\
\bottomrule
\end{tabular}
\end{sc}
\end{small}
\end{center}
\vskip -0.05in
\end{table}

\begin{table}[ht]
\caption{ResNet Hyperparameters}
\label{Resnet Final Hyperparameters}
\vskip 0.05 in
\begin{center}
\begin{small}
\begin{sc}
\begin{tabular}{lcccr}
\toprule
Hyperparameter &  Value\\
\midrule
Batch Size & 32 \\
Epochs  & 5 \\
Optimizer & Adam \\
Learning Rate & 0.001 \\
Loss Criterion   & Cross-Entropy \\
\bottomrule
\end{tabular}
\end{sc}
\end{small}
\end{center}
\vskip -0.05in
\end{table}

The best performing model based on test set accuracy, precision, recall is the fine-tuned ResNet-18. Interestingly, the results show that a deeper network doesn't necessarily translate to better model performance. While the ResNet-50 feature extraction model performed better than the ResNet-18 feature extraction model, the ResNet-18 fine-tuned model is superior to the ResNet-50 fine-tuned model. The ResNet-18 feature extraction model also performed better than both of the ResNet-50 models. This is probably due to the relatively small number of samples used to train the models. The deeper ResNet-50 architecture might be over-fitting or is not learning better representations of each image class but this is still speculated given the high complexity of ResNet-50.

\subsection{UNet with Dense Net}
\label{sec:UNet with Dense Net}
A semi-supervised approach can also be used to classify images, one of the more common methods is image segmentation. Image Segmentation is a method of image representation that uses a set of “masks” which act as a form of ground truth to segment specific portions of our images and these would be the representational patterns that we would like to capture. In this case, the pattern of interest is the rot present in an image, while identifying rot is a subjective matter it is entirely possible to map this as an input feature to a model and have its representations capture. This model would be able to capture image segments on related images. To achieve this, the model typically uses a combination of Double Convolutional Neural Networks with a structure called skip-connections which skip some of the connections in a neural network and feeds the output of one layer as input to the other layers. Skip-connections greatly reduce the complexity of loss surfaces, making it easier for optimizers to reduce loss while ensuring that feature representations are reused \cite{DBLP:journals/corr/abs-1712-09913}. The images for a sample image and prediction are shown below (areas in yellow are rotten areas of the meat as identified by the model).

\begin{figure}[ht]
    \centering
    \includegraphics[scale=0.12]{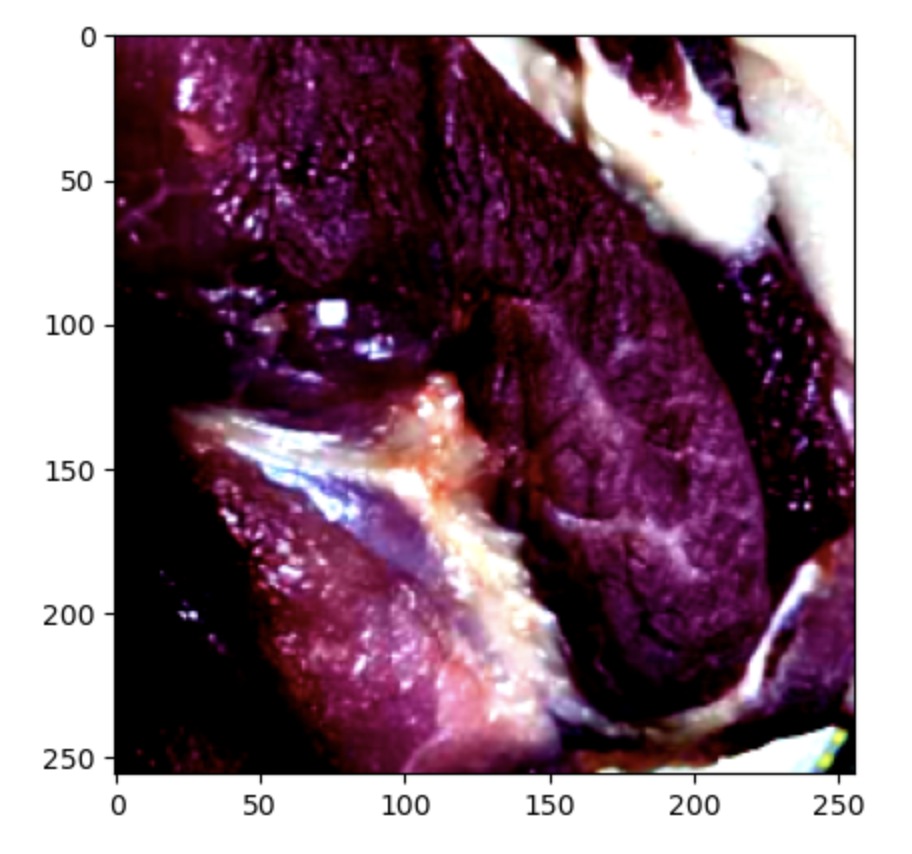}
    \includegraphics[scale=0.12]{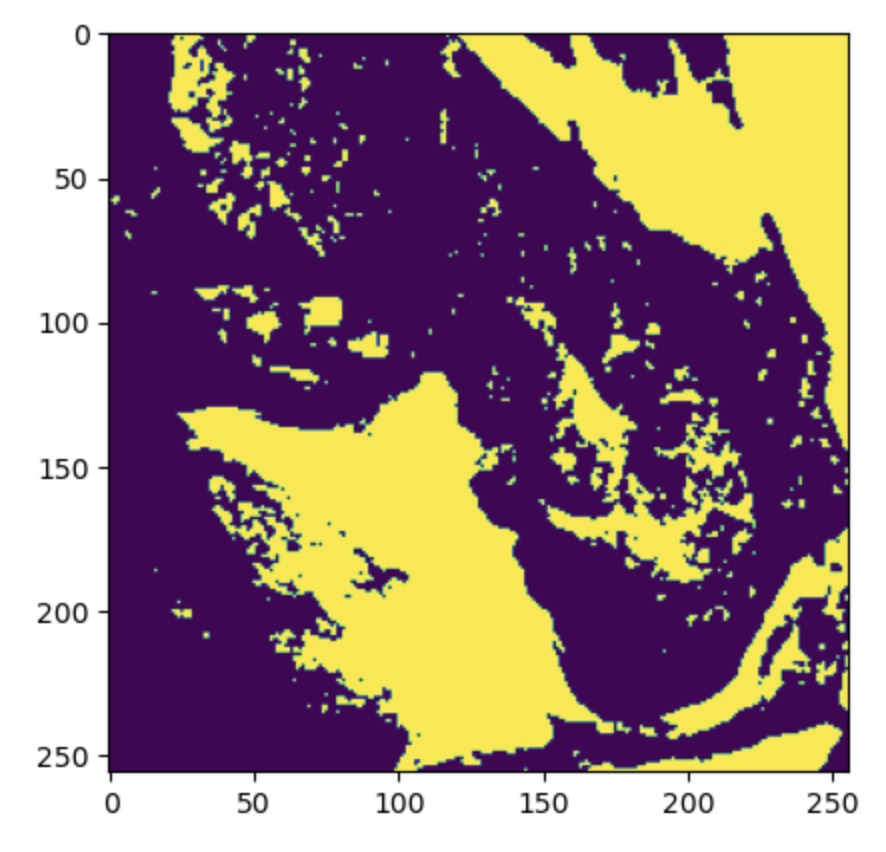}
    \caption{Sample Image and Prediction}
    \label{fig:Sample Image and Prediction}
\end{figure}

The algorithm used to achieve this was UNet \cite{DBLP:journals/corr/RonnebergerFB15}, it uses Double Convolutional layers to identify and extract features from the input image and uses skip connects to reuse these features in a related layer. The idea is that each feature set captured in a layer is captured in a layer connected by a skip connection and passed to the next layer to compute the representation segment. Since this task outputs a set of image patterns the ideal outcome would be identifying the quality of the outputs in terms of the intersection and the overlap resulting from the predictions and the image masks. The loss functions capable of representing this effort are Dice loss and Jaccard loss which broadly look at the ratio of the intersection to the union, so concretely both would have a measure of how well the model can segment the patterns of interest from a given input image.

The extracted image segmented predictions were passed as input features to a DenseNet model and predictions were output based on the segments captured. 

\begin{table}[ht]
\caption{UNet and DenseNet Hyperparameters}
\label{Unet Final Hyperparameters}
\begin{center}
\begin{small}
\begin{sc}
\begin{tabular}{lcccr}
\toprule
Hyperparameter &  Value\\
\midrule
Batch Size & 8 \\
Epochs  & 5 \\
Optimizer & AdamW \\
Learning Rate & 0.001 \\
Loss Criteria   & Jaccard Loss \& Cross-Entropy \\
\bottomrule
\end{tabular}
\end{sc}
\end{small}
\end{center}
\end{table}

The segment of interest in this case is the rot present in the image and this was one-hot encoded when it was passed to the DenseNet model. The outputs from this model would be used to classify the image.

\begin{table}[ht]
\caption{UNet with Dense Net Model Performance}
\centering
\label{UNet with Dense Net Model Performance}
\vskip 0.05 in
\begin{center}
\begin{small}
\begin{sc}
\begin{tabular}{lcccr}
\toprule
Accuracy & Precision & Recall \\
\midrule
35.25\% & 100\% & 35.25\% \\
\bottomrule
\end{tabular}
\end{sc}
\end{small}
\end{center}
\vskip -0.05in
\end{table}

This model, however, seems to show very poor performance in classification because the segments may not be fully interpreted by the model. While a larger model like ResNet learns several feature representations from an image with increasing complexity, this model learns only from the image segments captured and can only use this limited information for inference. As observed in the model performance, the recall is quite low which means that the model is incorrectly classifying meat but the precision is quite high which implies that the model is very accurate in identifying the correct classes. The resulting misclassification cost is also quite high in this case because every incorrect prediction would result in a very large cost to the business and as a result, this model was not used as the final model.

\section{Model Evaluation}
\label{Model Evaluation}
To evaluate the model performance, the primary metrics used were accuracy and Misclassification Cost (MCC). While accuracy is commonly understood as the model's predictive capability, another method of assessment would be MCC which uses the underlying principles of the Expected Value Framework (EVF) to identify the cost to a company based on the predictions from this model. MCC can be interpreted as the amount of money lost by the business should the model misclassifies an image. This would determine how the model can affect the business. 

The MCC is calculated based on an individual value resulting from the actual value vs the predictions as referenced in Table \ref{MCC} and \ref{Calculation of MCC} and cumulatively they would form a cost representing the amount of money lost by the business per misclassified image. This would prioritize the model development to ensure that specific costly misclassification, which is predicting actual spoiled as fresh or half-fresh, is avoided while ensuring that the model has a high accuracy. The cumulative MCCs shown in Table \ref{Evaluation on MCC} indicate the potential costs of using the model in one business day. It means that if daily benefits the client would obtain with this technology, such as labour cost reduction, outweight the cumulative MCC, the client could consider the introduction of the technology.

\begin{table}[ht]
\caption{Evaluation on MCC \\FE: Feature Extraction, FT: Fine-tuning}
\centering
\label{Evaluation on MCC}
\vskip 0.03 in
\begin{center}
\begin{small}
\begin{sc}
\begin{tabular}{lcccr}
\toprule
Model & Accuracy & MCC \\
\midrule
ResNet 18-FE & 82.70\% & \$886 \\
ResNet 18-FT  & \textbf{93.13\%} & \$5,076 \\
ResNet 50-FE   & 88.03\% & \textbf{\$242}  \\
ResNet 50-FT   & 84.70\% & \$316\\
UNet   & 35.25\% & \$89,411\\
\bottomrule
\end{tabular}
\end{sc}
\end{small}
\end{center}
\end{table}

\begin{figure}[!ht]
    \centering
    \includegraphics[scale=0.5]{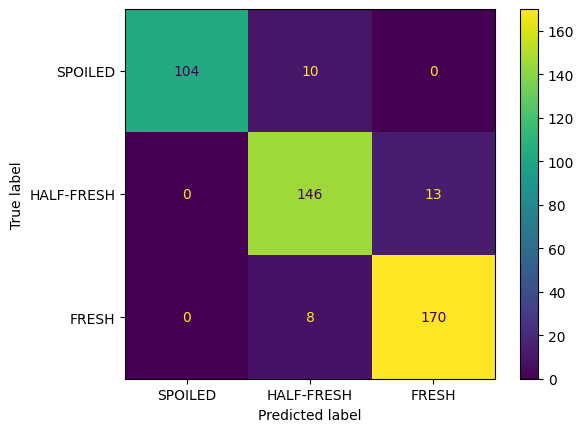}
    \caption{Confusion Matrix of 18FT}
    \label{fig:Confusion Matrix of 18FT}
\end{figure}
\newpage

\begin{figure}[!ht]
    \centering
    \includegraphics[scale=0.5]{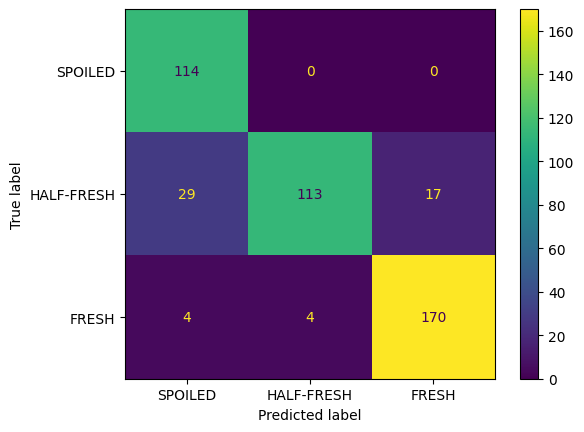}
    \caption{Confusion Matrix of 50FE}
    \label{fig:Confusion Matrix of 50FE}
\end{figure}

\begin{figure}[!ht]
    \centering
    \includegraphics[scale=0.3]{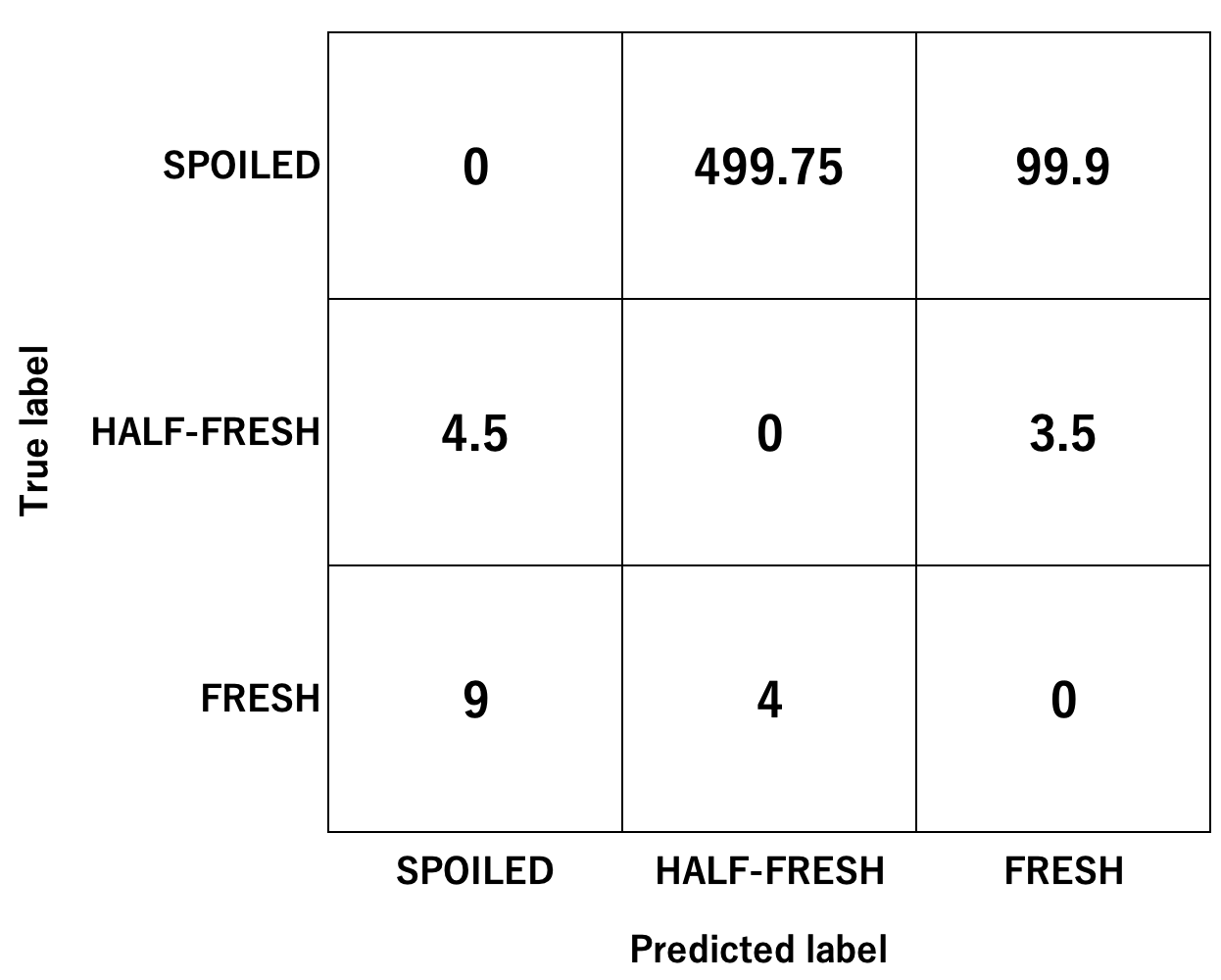}
    \caption{Matrix of MisClassification Cost(MCC)}
    \label{fig:Matrix of MCC}
\end{figure}

The ResNet 18-FT model has exceptionally higher precision and recall scores on test data compared to other models and this also shows that it is able to largely generalize on the dataset and given that the dataset is small it can identify the patterns correctly without compromising too much on the quality of the predictions. However, the ResNet 50-FE yielded the lowest MCC despite having lower accuracy, precision and recall compared to ResNet 18-FT. This means that ResNet 50-FE is the best model for this paper's business case. 

The reason why the MCC evaluation chose a lesser performing model as the most appropriate one for the business case lies in how the MCC matrix penalizes the mistakes of the models. Looking at the confusion matrix of ResNet 18-FT in Figure \ref{fig:Confusion Matrix of 18FT}, it misclassified 10 spoiled meat images as half-fresh, resulting in an MCC cost of \$4,998 on these mistakes alone. Contrast this with the ResNet 50-FE model in Figure \ref{fig:Confusion Matrix of 50FE}, where it didn't misclassify any spoiled meat images but made most of its mistakes misclassifying half-fresh meat images. The ResNet 50-FE model did not incur any heavy cost in misclassfying spoiled meat, and all of its other mistakes only incurred a cost of only \$242. This result is in line with the business case where selling a customer spoiled meat will incur a very a high cost, and consequently a model that misclassifies spoiled meat will incur a significant cost to the client.

In summary, it is recommended that the model to be deployed in production is the ResNet 50-FE model, since it will yield the client the lowest possible cost when this model makes mistakes.

\section{Interpretation}
SHAP \cite{SHAP} and LIME \cite{lime} paradigms were used to understand how the model works and improve interpretability of the model, to identify what features or areas of the image the model uses to identify the class of a particular piece of meat. Results from SHAP were inconclusive and ambiguous, however, results of using LIME offered valuable insight into what the model sees and uses to perform classification. Some results from the LIME classification are given below.

\label{Interpretation}
\begin{figure}[ht]
    \centering
    \includegraphics[scale=0.28]{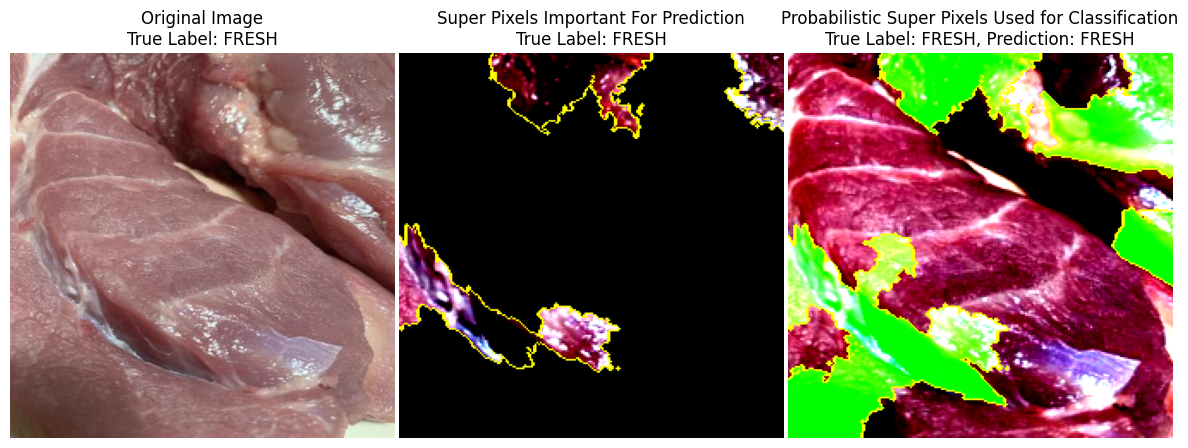}
    \vskip 0.1in
    \includegraphics[scale=0.28]{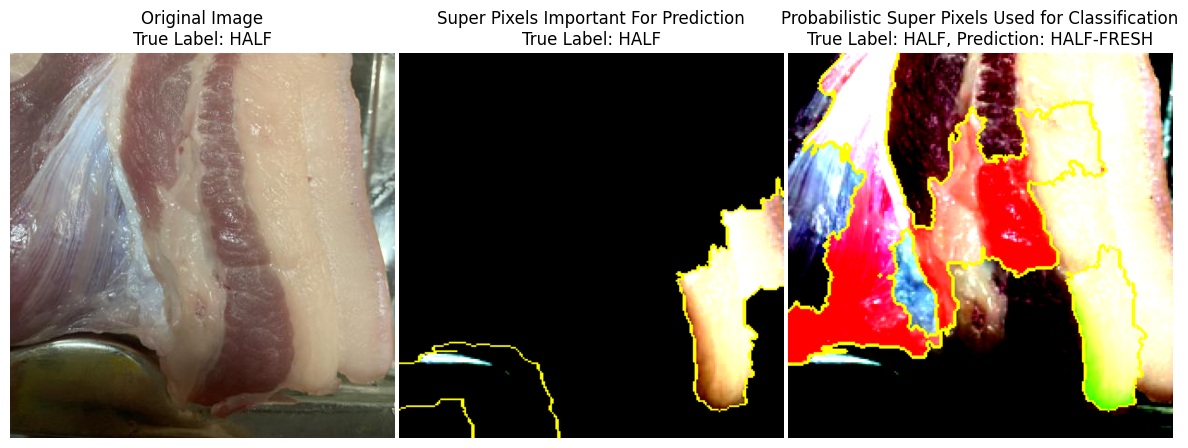}
    \vskip 0.1in
    \includegraphics[scale=0.28]{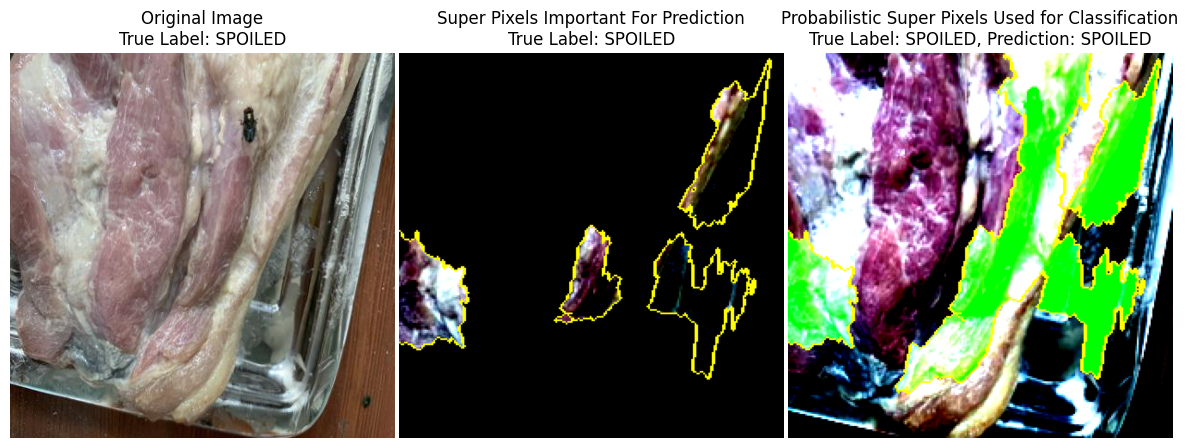}
    \caption{Example of Lime Interpretations for each of three classes: Fresh (Top), Half-Fresh (Middle) and Spoiled (Bottom)}
    \label{fig:Example of Lime Interpretationsfor each of three classes: Fresh (Top), Half-Fresh (Middle) and Spoiled (Bottom)}
\end{figure}

The images in the middle represent the super pixels or segments used as important features used by the model to classify the image for a specific class and the images on the right represent the probabilistic regions used by the model for classification; with regions in green indicating a higher probability that the model used those regions while regions in red indicating a lower probability for the same. 

It is observed that the model is able to identify the key regions of rot in the spoilt meat and use those regions for determining it's classification. On the other hand, the model is also able to identify similar segments of freshness in fresh meet and classify those correctly as well. This suggests that the model developed is able to differentiate between spurious features in the image and pick out the important segments of the image that will help it in classification. 

\begin{figure}[ht]
    \centering
    \includegraphics[scale=0.28]{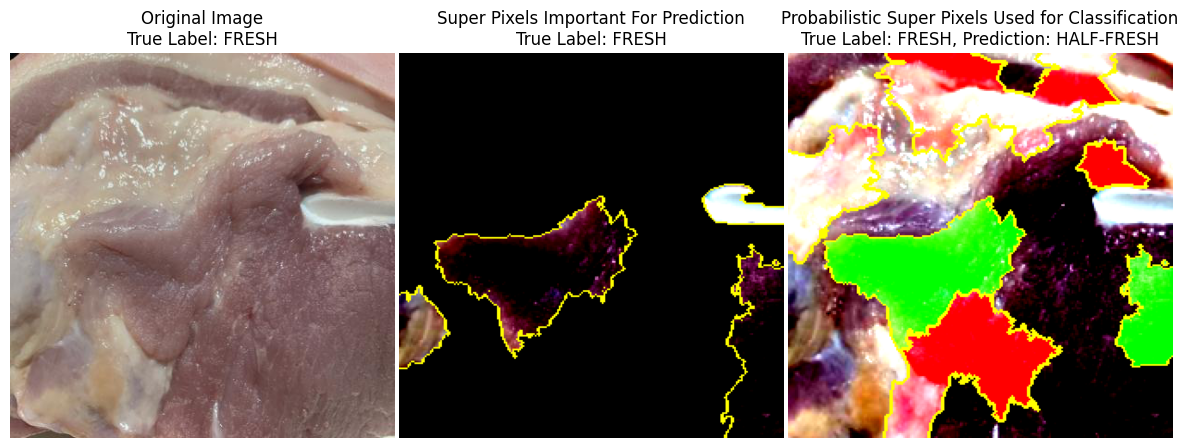}
    \vskip 0.1in
    \includegraphics[scale=0.28]{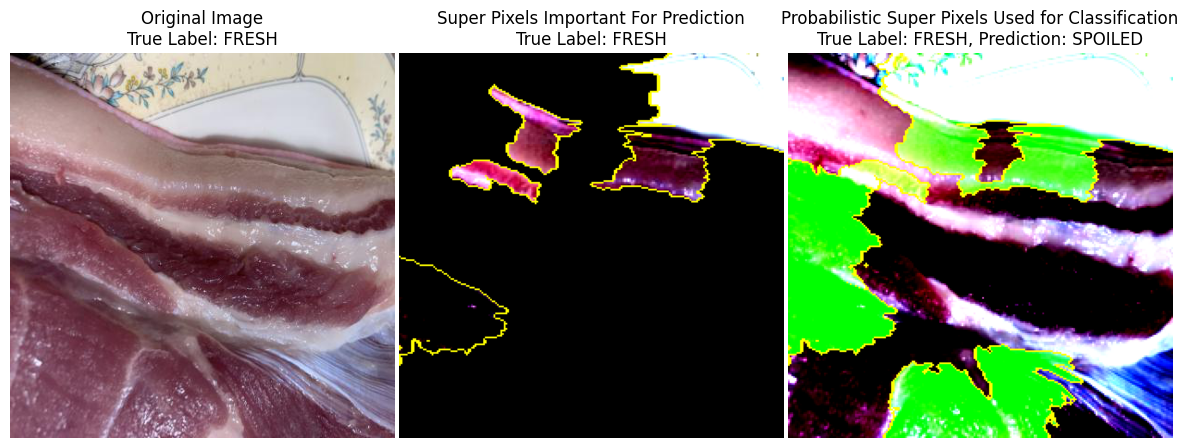}
    \vskip 0.1in
    \includegraphics[scale=0.28]{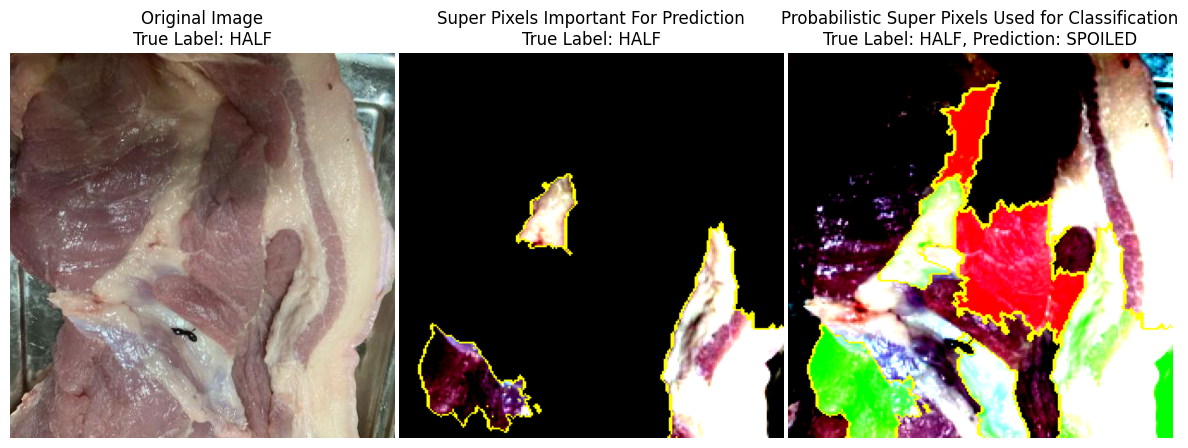}
    \vskip 0.1in
    \includegraphics[scale=0.28]{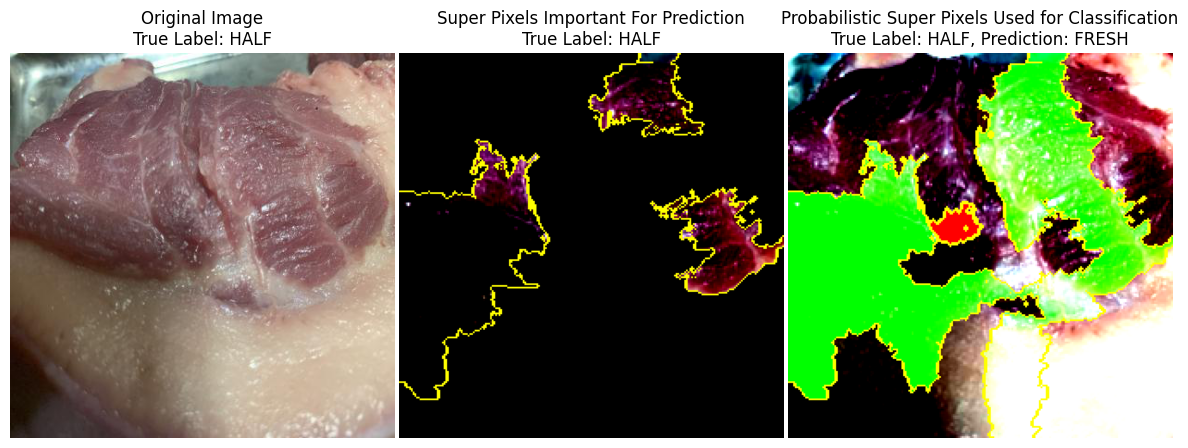}
    \caption{Example of Lime Interpretations for images that were misclassified}
    \label{fig:Examples of Lime Interpretations for images that were misclassified}
\end{figure}

A similar analysis of a few misclassified images (\ref{fig:Examples of Lime Interpretations for images that were misclassified}) suggests the same. Though the model is unable to correctly classify these food items, it is still successful in  identifying appropriate areas of the image which can serve as important input features in the final decision. 

It can hence be concluded that while the model does not yield 100\% accuracy, it's current decision making is based on identifying valid areas of the image that represent fresh or spoiled meat, rather than using spurious areas such as portion of packaging or image background to determine the same. 

\label{Misclassified Examples}

\section{Conclusion}
\label{Conclusion}
By utilizing this technology, clients would be able to enjoy three main benefits. The first is improved efficiency. By judging freshness based on the actual condition of the product rather than the number of days since processing, unnecessary discounts and waste can be avoided. Secondly, there is an improvement in reliability. Just because the number of days since processing is short does not necessarily mean that the product has not spoiled. This technology can identify truly spoiled products in real time to dispose of them. Avoiding the sale of defective products leads not only to maintaining customer trust but also to avoiding both monetary and non-monetary costs associated with consumers’ health problems due to the sale and consumption of spoiled food, which is non-trivial. The third is contributing to better corporate image. Recently, companies' contribution toward sustainability is being highlighted. If food waste can be reduced by this technology, it can be a marketing advantage and can contribute to an increase in corporate value.

However, there are also barriers to overcome in order to create the above-mentioned value. If the applicable ingredients are limited (such as only meat), supermarkets will not be interested in this technology. To introduce it into actual operation, it would be necessary to be applicable to all kinds of perishable foods. Also, freshness standards may differ depending on the weather at that time. To address these issues, a huge amount of training data and time, as well as new features to consider additional factors such as humidity levels, are needed. In addition, this model assumes that clear images are available for each slice of meat. In other words, it assumes that each product on display can be photographed one by one with adequate lighting and that the meat is not blocked by any packaging or other material. If a model is improved with cameras and object identification/image processing such that it can analyze the freshness of multiple products at the same time from a single image containing multiple products with packaging, the usability of this model will be further enhanced.

Additionally, the interpretation portrayed using LIME can be used in multiple ways. At any given point, it can be used to generate a similar probability map and check what are the areas of rot on the meat that the model is using to make it's prediction. This can help understand if the model is 'seeing' the correct features. This utility can be further extended to monitor model performance, and track deterioration if the model starts using irrelevant or relatively unimportant areas of the image to make its classification. Any deterioration or change in the probability maps would signal a need to retrain the model or deploy new models

In this project, a supermarket is assumed as a client. However, this technology can also have other applications. For consumers, if a device that can detect the food conditions can be installed in their refrigerators, they can reduce food expenses and waste at home.

Finally, the utility and value of this model can be further enhanced by merging it with data-driven decisions such as maintaining inventory based on demand forecasting and other business analytics techniques to add multiple layers of safety in terms of food freshness and wastage.

\bibliography{main.bib}

\subsection*{Reproducibility}

The source code and data used to generate the results presented in this paper are available at: \href{https://github.com/TheLohia/Phteven}{\texttt{https://github.com/TheLohia/Phteven}}. This contains: (i) Model training and evaluations scripts (ii) Jupyter notebooks for model experiments and (iii) web demo for the model

\end{document}